\documentclass[sigconf]{acmart}

\acmConference{EARL 25: The 2nd Workshop on Evaluating and Applying Recommender Systems with Large Language Models at RecSys '25}{Sept 22--26, 2025}{Prague, Czech Republic}
\acmYear{2025}

\acmBooktitle{EARL '25 at RecSys '25, Sept 22--26, 2025, Prague, Czech Republic}

\renewcommand\footnotetextcopyrightpermission[1]{}

\usepackage{graphicx} 
\usepackage{makecell}
\usepackage{colortbl}
\usepackage{pifont}

\usepackage[skip=1ex]{parskip}  
\usepackage{booktabs}
\usepackage{siunitx}
\usepackage{subcaption}
\usepackage{tabularx}
\usepackage{tcolorbox} 
\usepackage{placeins} 

\usepackage{csquotes}         
\usepackage[capitalize,nameinlink,noabbrev]{cleveref}
\usepackage{balance}

\makeatletter
\def\fps@table{htbp!}                 
\def\fps@figure{htbp!}                
\makeatother

\title[LLM-as-a-Judge: Rapid Evaluation of Legal Document Recommendation for Retrieval-Augmented Generation]{LLM-as-a-Judge: Rapid Evaluation of Legal Document Recommendation for Retrieval-Augmented Generation}

\author{Anu Pradhan, Alexandra Ortan, \\ Apurv Verma, Madhavan Seshadri}

\affiliation{%
  \institution{Bloomberg}
  \city{New York}
  \state{NY}
  \country{USA}
}
\email{{apradhan11, aortan, averma239, mseshadri}@bloomberg.net}

\begin{document}

\renewcommand{\shortauthors}{Pradhan et al.}

\begin{abstract}
The evaluation bottleneck in recommendation systems has become particularly acute with the rise of Generative AI, where traditional metrics fall short of capturing nuanced quality dimensions that matter in specialized domains like legal research. Can we trust Large Language Models to serve as reliable judges of their own kind? This paper investigates LLM-as-a-Judge as a principled approach to evaluating Retrieval-Augmented Generation systems in legal contexts, where the stakes of recommendation quality are exceptionally high.

We tackle two fundamental questions that determine practical viability: which inter-rater reliability metrics best capture the alignment between LLM and human assessments, and how do we conduct statistically sound comparisons between competing systems? Through systematic experimentation, we discover that traditional agreement metrics like Krippendorff's alpha can be misleading in the skewed distributions typical of AI system evaluations. Instead, Gwet's AC2 and rank correlation coefficients emerge as more robust indicators for judge selection, while the Wilcoxon Signed-Rank Test with Benjamini-Hochberg corrections provides the statistical rigor needed for reliable system comparisons.

Our findings suggest a path toward scalable, cost-effective evaluation that maintains the precision demanded by legal applications—transforming what was once a human-intensive bottleneck into an automated, yet statistically principled, evaluation framework.

\end{abstract}
\keywords{LLM-as-a-Judge, Large Language Models, RAG, Evaluation}
\maketitle
\makeatletter
\fancyhead[RO,RE]{}
\makeatother
\section{Introduction}
Recommendation systems play an essential role across various sectors, including legal research, where the accuracy and quality of recommended documents can directly impact professional decision-making. The advent of LLMs has revolutionized the field of natural language processing, offering unprecedented capabilities in understanding and generating human-like text. These models, such as GPT-4 and others, have shown promise not only in generating content but also in evaluating and assessing - a concept we refer to as "LLM-as-a-Judge". Leveraging LLMs in evaluation tasks holds significant potential for increasing efficiency and consistency in areas like search relevancy and answer quality assessments. However, using LLMs as evaluators introduces challenges related to the reliability and variability of their judgments compared to human raters.\par

We view the LLM-as-a-Judge as an artificial rater, and explore the use of Inter-Rater Reliability (IRR) metrics to measure the reliability with human raters. IRR is the degree of agreement among different raters assessing the same set of items. High IRR indicates consistent and reliable ratings, which are essential for tasks like tuning search algorithms or training machine learning models. 
In this paper, we explore the application of LLM-as-a-Judge to evaluate different generative AI  (Gen AI) solutions across different Bloomberg Law products. In particular, we study how multiple raters—both human and LLM-based—evaluate items on ordinal scales assessing relevance, safety, hallucinations, correctness, and overall quality. We formulate two research questions (RQs) to guide our investigation.

\begin{table*}[htbp]
\centering
\small
\renewcommand{\arraystretch}{1.3}
\setlength{\tabcolsep}{6pt}
\begin{tabular}{@{}l*{7}{c}@{}}
\toprule
\cellcolor{gray!10}\textbf{IRR Metric} &
\cellcolor{gray!10}\makecell{\textbf{Ordinal}\\\textbf{Scale}\\\textbf{Support}} &
\cellcolor{gray!10}\makecell{\textbf{Distribution}\\\textbf{Robustness}} &
\cellcolor{gray!10}\makecell{\textbf{Rank}\\\textbf{Sensitivity}} &
\cellcolor{gray!10}\makecell{\textbf{Clear}\\\textbf{Interpretation}} &
\cellcolor{gray!10}\makecell{\textbf{Missing}\\\textbf{Data}\\\textbf{Tolerance}} &
\cellcolor{gray!10}\makecell{\textbf{Computational}\\\textbf{Efficiency}} &
\cellcolor{gray!10}\makecell{\textbf{Multi-Rater}\\\textbf{Capability}} \\
\midrule
Cohen's Kappa & {\color{red}\ding{55}} & {\color{red}\ding{55}} & {\color{red}\ding{55}} & {\color{green!60!black}\ding{51}} & {\color{red}\ding{55}} & {\color{green!60!black}\ding{51}} & {\color{red}\ding{55}} \\
Percent Agreement & {\color{red}\ding{55}} & {\color{red}\ding{55}} & {\color{red}\ding{55}} & {\color{green!60!black}\ding{51}} & {\color{red}\ding{55}} & {\color{green!60!black}\ding{51}} & {\color{green!60!black}\ding{51}} \\
Kendall's Tau & {\color{green!60!black}\ding{51}} & {\color{green!60!black}\ding{51}} & {\color{green!60!black}\ding{51}} & {\color{green!60!black}\ding{51}} & {\color{red}\ding{55}} & {\color{red}\ding{55}} & {\color{red}\ding{55}} \\
Spearman's Rank & {\color{green!60!black}\ding{51}} & {\color{green!60!black}\ding{51}} & {\color{green!60!black}\ding{51}} & {\color{green!60!black}\ding{51}} & {\color{red}\ding{55}} & {\color{green!60!black}\ding{51}} & {\color{red}\ding{55}} \\
Krippendorff's Alpha & {\color{green!60!black}\ding{51}} & {\color{red}\ding{55}} & {\color{green!60!black}\ding{51}} & {\color{green!60!black}\ding{51}} & {\color{green!60!black}\ding{51}} & {\color{green!60!black}\ding{51}} & {\color{green!60!black}\ding{51}} \\
Gwet's AC2 & {\color{green!60!black}\ding{51}} & {\color{green!60!black}\ding{51}} & {\color{green!60!black}\ding{51}} & {\color{green!60!black}\ding{51}} & {\color{green!60!black}\ding{51}} & {\color{green!60!black}\ding{51}} & {\color{green!60!black}\ding{51}} \\
\bottomrule
\end{tabular}
\caption{Comparison of IRR Metrics for Legal RAG System Evaluation. \textbf{Note:} {\color{green!60!black}\ding{51}} indicates the metric supports the attribute; {\color{red}\ding{55}} indicates it does not.}
\label{tab:irr-metrics}
\end{table*}

\textbf{RQ1: How can we effectively evaluate and select LLM judges for legal RAG systems using a comprehensive set of inter-rater reliability metrics?} To address the question of how to effectively evaluate and select LLM judges for legal RAG systems, we emphasize the need to consider a comprehensive set of inter-rater reliability metrics rather than relying on a single measure. This multimetric approach is crucial due to the complex nature of legal language and the varied challenges in AI evaluation. We examine traditional IRR metrics like Krippendorff's Alpha (K-Alpha), which, while widely used, may misrepresent agreement levels in skewed data distributions common in Gen AI system evaluations. To address these limitations, we also explore more recent techniques such as Gwet's AC2, which offers improved robustness in these scenarios. Additionally, we consider correlation metrics such as Spearman rank correlation and Kendall tau ($\tau$), particularly suited for ordinal ratings and measuring agreement on relative rankings \citep{spearman, kendal-tau, Gwet2008ComputingIR}. By evaluating this diverse set of metrics based on their ability to measure LLM-human agreement, handle skewed distributions, and reliably rank Gen AI systems, our aim is to provide a nuanced and comprehensive framework for selecting LLM judges. 

\textbf{RQ2: Which statistical methods are most effective for comparing legal RAG systems evaluated by LLM judges, and how do different multiple hypothesis testing corrections impact these comparisons?} In evaluating LLM judges for legal RAG systems, we carefully selected the Wilcoxon Signed-Rank Test (WSRT) as our primary statistical method due to its nonparametric nature, which is ideal for ordinal data from LLM evaluations that may not follow normal distributions \citep{wilcoxon}. To address multiple testing issues, we evaluated three correction methods: Bonferroni, Benjamini-Hochberg (B-H), and Holm-Bonferroni. Given the exploratory nature of our study and the multiple performance aspects, we selected the B-H method to balance error control with detection sensitivity \citep{chen_2017}. This approach allows us to effectively identify significant differences in various legal comparisons of the RAG system while maintaining statistical rigor.

\section{Related Work}
Retrieval Augmented Generation (RAG) effectively enhances the capabilities of LLMs by integrating external knowledge sources, making it valuable for recommendation tasks in specialized domains \citep{lewis_retrieval-augmented_2020, gao2024retrievalaugmentedgenerationlargelanguage}. Traditional automated evaluation metrics such as ROUGE and BLEU depend on reference responses, which limits their effectiveness in complex open-ended recommendation scenarios \citep{lin-2004-rouge, 10.3115/1073083.1073135, zhang2020bertscoreevaluatingtextgeneration}. While human evaluations are typically considered the gold standard due to their accuracy, they are impractical at large scales given the significant time and expertise required \citep{doostmohammadi2024reliableautomaticevaluationmethods}. Recent advances in LLMs have sparked interest in their potential as automated evaluators, particularly in specialized domains, such as law. This section examines three key areas of relevant literature: the use of LLMs as judges, challenges in ensuring reliable evaluations, and statistical methods for analyzing LLM-based assessments.

\textbf{LLM-as-a-Judge:} Using LLMs as automated judges has emerged as a scalable alternative to human evaluation \citep{10.5555/3666122.3668142,liu-etal-2023-g}. Although GPT-4 has shown promise in achieving human-level agreement on certain tasks \citep{sottana-etal-2023-evaluation}, recent studies have identified key challenges, including cognitive biases \citep{koo-etal-2024-benchmarking}, self-preference \citep{Panickssery2024LLMER}, and systematic errors in evaluation \citep{wang-etal-2024-large-language-models-fair}. To address these limitations, researchers have explored committee-based approaches using multiple LLM \citep{Verga2024ReplacingJW,Chan2023ChatEvalTB} and specialized training of smaller expert judges \citep{kim-etal-2024-prometheus,zhu2023judgelm}. However, these methods often lack rigorous guarantees of reliability and agreement with human preferences.

\textbf{Reliable Evaluation:} While LLMs can scale evaluation across large datasets, ensuring reliability as judges presents significant challenges in terms of cognitive biases \citep{koo-etal-2024-benchmarking}, self-preference \citep{Panickssery2024LLMER} and systematic errors in evaluation \citep{wang-etal-2024-large-language-models-fair}. To address these challenges, researchers have explored various approaches.  For instance, Chan et al. (2023) proposed a multi-agent debate framework to mitigate individual model biases \citep{Chan2023ChatEvalTB}. Sottana et al. (2023) demonstrated that GPT-4 can achieve human-level agreement on certain tasks, but noted persistent challenges in complex evaluations \citep{sottana-etal-2023-evaluation}. These studies collectively underscore the importance of developing robust methods for LLM-based evaluations.

\textbf{Statistical Analysis: } Recent studies have emphasized the critical need for robust statistical methods to evaluate language models, highlighting issues in experimental design, variance quantification and uncertainty estimation. Miller (2024) introduced a framework for adding error bars to LLM evaluations, proposing the use of paired differences at the question level for statistical inference \citep{miller2024addingerrorbarsevals}. Oosterhuis et al. (2024) developed methods to construct reliable confidence intervals for IR evaluation metrics using LLM-generated annotations \citep{10.1145/3637528.3671883}. Card et al. (2020) revealed the prevalence of underpowered experiments in NLP, particularly in popular benchmarks, highlighting the importance of power analysis in experimental design \citep{card-etal-2020-little}. Together, these works offer valuable information for designing statistically sound experiments and reliable evaluation procedures for LLMs.
\section{Experimental Design}
We evaluated several metrics, including Cohen's Kappa \citep{cohen1960coefficient}, Krippendorff's Alpha \citep{krippendorff2018content}, Spearman's rank correlation \citep{spearman1961proof}, Kendall's Tau \citep{kendall1938new}, percent agreement and two extensions of Gwet's AC2 with linear weighting schemes (Gwet AC2-L) and quadratic (Gwet AC2-Q) \citep{gwet2008computing}. While we did not evaluate every available IRR metric, such as Fleiss's Kappa \citep{fleiss1971measuring} and the Brennan-Prediger coefficient \citep{brennan1981coefficient}, our aim was to emphasize the importance of utilizing a diverse set of metrics rather than conducting an exhaustive analysis of all available IRR metrics.e evaluated several metrics, including Cohen's Kappa \citep{cohen1960coefficient}, Krippendorff's Alpha \citep{krippendorff2018content}, Spearman's rank correlation, Kendall's Tau, percent agreement and two extensions of Gwet's AC2 with linear weighting schemes (Gwet AC2-L) and quadratic (Gwet AC2-Q). While we did not evaluate every available IRR metric, such as Fleiss's Kappa and the Brennan-Prediger coefficient, our aim was to emphasize the importance of utilizing a diverse set of metrics rather than conducting an exhaustive analysis of all available IRR metrics. As a leading provider of legal information services, Bloomberg Law faces unique challenges in evaluating RAG systems. Legal content requires high accuracy standards, as even minor errors could impact critical legal decisions. Additionally, the proprietary nature of our content and systems, combined with client confidentiality requirements, creates constraints on sharing detailed system specifications or complete evaluation results. Our evaluation was conducted on two legal RAG systems operating over a comprehensive legal corpus reflecting real production challenges at Bloomberg Law where we evaluate thousands of legal query-document pairs monthly across multiple products. Each RAG system consists of two critical components: a retrieval component that identifies relevant legal documents, and an answer generation component that synthesizes the retrieved information into coherent responses.

\begin{table*}[!t]
\centering
\small
\renewcommand{\arraystretch}{1.3}
\setlength{\tabcolsep}{8pt}
\begin{tabular}{@{}l*{4}{c}@{}}
\toprule
\cellcolor{gray!10}\textbf{Attribute} & \cellcolor{gray!10}\textbf{Wilcoxon} & \cellcolor{gray!10}\textbf{Sign Test} & \cellcolor{gray!10}\textbf{Mann\textendash Whitney} & \cellcolor{gray!10}\textbf{Friedman} \\
\midrule
Statistical Power & \textsc{High} & \textsc{Moderate} & \textsc{High} & \textsc{Moderate} \\
Sample Size Needed & \textsc{Small} & \textsc{Small} & \textsc{Medium} & \textsc{Medium} \\
Study Design & \textsc{Paired} & \textsc{Paired} & \textsc{Independent} & \textsc{Repeated} \\
Robustness & \textsc{High} & \textsc{Very High} & \textsc{High} & \textsc{High} \\
\midrule
\cellcolor{blue!5}\textbf{When to Use:} & \cellcolor{blue!5} & \cellcolor{blue!5} & \cellcolor{blue!5} & \cellcolor{blue!5} \\
Best For & \makecell{A vs B comparison\\ \textit{(magnitude matters)}} & \makecell{A vs B comparison\\ \textit{(simple win/loss)}} & \makecell{Two groups\\ \textit{(independent data)}} & \makecell{Multiple systems\\ \textit{(3+ comparisons)}} \\
\bottomrule
\end{tabular}
\caption{Statistical test comparison for LLM evaluation. \textbf{Our choice:} Wilcoxon for paired A/B testing with magnitude sensitivity.}
\label{tab:nonparametric_tests}
\end{table*}

We compared two distinct legal RAG systems. System A utilizes traditional BM25 retrieval combined with an open-source LLM summarizer applied to the top 5 retrieved documents. System B incorporates improvements in the retrieval system and employs the proprietary GPT-4 model by OpenAI as the summarizer. This comparison reflects realistic industry scenarios and evaluates significant technological enhancements and their practical impacts
on recommendation quality. Our evaluation framework specifically targeted both these components: the retrieval effectiveness through search relevancy assessment, and the generation quality through answer evaluation. For search relevancy, we evaluated passage-query pair relevance on a scale of 1 to 4, while the answer quality assessment examined multiple dimensions including relevance, conciseness, readability, completeness, and extrinsic hallucination using the same scale of 1 to 4, while the answer quality assessment examined multiple dimensions including relevance, conciseness, readability, completeness, and extrinsic hallucination using the same scale \citep{puglia2024}.

The evaluation dataset consisted of 117 anonymized legal user queries, carefully selected to represent actual user interactions, including legal research queries from practicing attorneys and librarians. Although generating large-scale, expert-curated queries poses practical challenges due to the significant domain expertise and effort required, our dataset size is realistic and practically representative of typical industry evaluations. Additionally, the selected statistical methods, particularly non-parametric tests such as the WSRT with BH corrections, are robust and specifically suitable for datasets of this scale, ensuring the statistical validity and reliability of our findings. For each query, both systems generated answers in the form of summaries derived from the top-k retrieved documents, accompanied by supporting references. The answers were structured to provide concise legal analyses while maintaining traceability to source documents.

\subsection{IRR Metric Analysis}
In this study, we investigated and calculated various IRR metrics in human and LLM evaluation data sets (see \autoref{tab:irr-metrics}). Our goal is to discover a metric or combination of metrics that can guide us in selecting the most suitable LLM model for specific evaluation tasks. Currently, numerous proprietary and open-source LLM models are available for such tasks. Evaluation tasks often involve different measurement levels, including nominal (e.g., categories), ordinal (e.g., rankings), and, to a lesser extent, interval and ratio data. Most evaluations focus on nominal and ordinal data, underscoring the importance of identifying a metric set tailored to these tasks. In addition, IRR metrics typically assume specific distributions of categories and ratings to calculate the percentage of chance agreement. When this assumption is violated, the metrics may not provide the expected insight. For instance, in situations with skewed ratings distributions, relying solely on Krippendorff's Alpha \citep{krippendorff2018content} might lead to an underestimation of raters' agreement due to inflated chance agreement calculations. Gwet's AC2 offers a solution by providing a more stable estimation that is less affected by category prevalence, thus giving a truer reflection of IRR.

We evaluated several metrics, including Cohen's Kappa, Krippendorff's Alpha, Spearman's rank correlation, Kendall's Tau, percent agreement and two extensions of Gwet’s AC2 with linear weighting schemes (Gwet AC2-L) and quadratic (Gwet AC2-Q). While we did not evaluate every available IRR metric, such as Fleiss's Kappa and the Brennan-Prediger coefficient, our aim was to emphasize the importance of utilizing a diverse set of metrics rather than conducting an exhaustive analysis of all available IRR metrics.

\subsection{Statistical Testing}
In the context of evaluating LLM outputs for legal RAG systems, we carefully selected five key attributes to compare various nonparametric tests. Relative power, impact of sample size, number of repeated measurements, robustness and practical implications (see \autoref{tab:nonparametric_tests}). \textbf{Relative Power} was chosen for its critical role in detecting subtle differences between LLM judges or RAG systems, which is essential given the nuanced nature of legal language and the potentially small but significant variations in system performance. \textbf{Sample Size Impact} was included due to practical constraints often faced in the legal evaluation of AI, where large datasets may not always be available or feasible to process. \textbf{Number of Repeated Measurements} attribute is crucial for aligning the statistical test with our experimental design, which typically involves comparing two systems or versions on the same set of legal queries. \textbf{Robustness} was selected to ensure the reliability of our results under various data conditions, acknowledging the diverse and sometimes unpredictable nature of legal text data. Finally, we included \textbf{Practical Implications} to provide context on each test's applicability, helping to bridge the gap between statistical theory and the practical challenges of evaluating legal AI systems. Based on these attributes, we determined that the Wilcoxon Signed Rank test \citep{wilcoxon1945individual} is the most appropriate nonparametric test for our specific context of LLM judge evaluation in legal RAG systems (see \autoref{tab:nonparametric_tests}).

This test compares two related samples to assess the differences in population mean rank. Calculate the differences between the paired observations, rank them, and derive a test statistic by summing the positive and negative ranks separately. Its importance lies in its ability to analyze nonnormally distributed data, making it a robust alternative to the paired t-test.  Additionally, we considered three multiple testing correction methods: Bonferroni \citep{bonferroni1936teoria}, Benjamini-Hochberg (B-H) \citep{benjamini1995controlling}, and Holm-Bonferroni \citep{holm1979simple}. Each balances strictness of correction, error control, and statistical power differently. Bonferroni, the most conservative, offers strong family-wise error rate (FWER) control but lower power, suitable for critical decisions with few comparisons. B-H controls the false discovery rate (FDR) with higher power, which is ideal for exploratory analyses with numerous comparisons. Holm-Bonferroni provides a middle ground. Given our study's exploratory nature and multiple performance aspects, we selected the B-H method to balance error control and detection of significant differences.

\begin{table}[H]
	\centering
	\small
	\begin{tabular}{l r}
		\toprule
		\rowcolor{gray!20}
		{\textbf{Metric}} & {\textbf{Skewness}} \\
		\midrule
		Relevance                 & $-0.4895$ \\
		Completeness              & $ 1.0569$ \\
		Extrinsic Hallucinations  & $ 3.7119$ \\
		Readability               & $ 2.1468$ \\
		Correctness               & $ 0.0898$ \\
		Inaccurate Hallucinations & $ 5.1269$ \\
	\bottomrule
	\end{tabular}
	\caption{Distribution skewness of evaluation metrics used in RQ2. The strong right skew in hallucination-related metrics motivates non-parametric tests.}
	\label{tab:metric_skewness}
\end{table}

We used the Wilcoxon signed-rank test with B-H correction to compare two systems (A and B) using various metrics on 117 queries of varying complexity.
Using GPT-4o as a judge, we combined direct and pairwise assessments, performing 10 runs per query and taking the majority vote. This approach ensures rigor in identifying significant performance differences while mitigating false positives from multiple comparisons.

\section{Results}
Our analysis focused on two key aspects: evaluating different LLM judges for their reliability in assessing legal RAG systems (RQ1) and comparing the performance of two RAG systems using statistical methods (RQ2). For RQ1, we examined various inter-rater reliability metrics to determine their effectiveness in selecting appropriate LLM judges. For RQ2, we investigated the application of statistical methods, particularly the WSRT with B-H corrections, to compare system performance across multiple metrics.

\begin{table*}[!t]
	\centering
	\small
	\begin{tabular}{>{\hspace{0.4em}}l *{7}{S[table-format=1.2]}}
		\toprule
		\rowcolor{gray!20}
		{\textbf{LLM Judge}}
			& {\textbf{Percent Agr.}}
			& {\textbf{Cohen\,\(\kappa\)}}
			& {\textbf{Krippendorff\,\(\alpha\)}}
			& {\textbf{Gwet’s AC2 (lin)}}
			& {\textbf{Gwet’s AC2 (quad)}}
			& {\textbf{Spearman}}
			& {\textbf{Kendall \(\tau\)}} \\
		\midrule
GPT4o            & {\bfseries 0.56} & {\bfseries 0.35} & {\bfseries 0.70} & {\bfseries 0.63} & {\bfseries 0.78} & {\bfseries 0.73} & {\bfseries 0.66} \\
LLaMA2-70B       & 0.26 & 0.07 & 0.32 & 0.26 & 0.47 & 0.68 & 0.61 \\
Claude Opus3     & 0.40 & 0.21 & 0.43 & 0.35 & 0.48 & 0.64 & 0.56 \\
Mistral          & 0.42 & 0.21 & 0.52 & 0.50 & 0.70 & 0.64 & 0.57 \\
Prometheus2-7B   & 0.30 & 0.13 & 0.06 & 0.17 & 0.27 & 0.42 & 0.37 \\
Prometheus2-8x7B & 0.40 & 0.17 & 0.43 & 0.36 & 0.46 & 0.53 & 0.46 \\
	\bottomrule
	\end{tabular}
	\caption{LLM judge agreement and ranking consistency on Search Relevancy. Best per column in bold. \(\kappa\): Cohen’s kappa; \(\alpha\): Krippendorff’s alpha. This supports the RQ1 findings discussed in the text.}
	\label{tab:llm_comparison}
\end{table*}

\subsection{RQ1 Finding: Evaluating LLM Judges}

Our analysis of IRR metrics reveals specific recommendations for different evaluation scenarios in the legal domain. For general agreement assessment, K-Alpha proves effective with balanced rating distributions, while Gwet's AC2 is preferred for skewed distributions (\autoref{tab:llm_comparison}). The correlation metrics in our study specifically measure different aspects of ranking consistency. Spearman's rank correlation evaluates how well LLM judges preserve the relative ordering of document relevance compared to human expert rankings (with GPT4o showing the highest correlation at 0.73), while Kendall's Tau measures pairwise ranking consistency, particularly important for maintaining proper precedential value ordering between documents. For specific tasks, Gwet's AC2 with quadratic weighting demonstrated superior performance (0.78 for GPT4o) in assessing relevance, while Cohen's Kappa remains adequate for binary decisions despite its limitations with skewed data.

When applying these metrics to evaluate different LLM judges, we observed varying strengths and weaknesses that highlight the importance of a multimetric approach. For example, Prometheus2 8x7B exhibits a higher K-Alpha (0.43) than the Llama model (0.32), suggesting a better overall agreement. However, for ordinal data where rank preservation is crucial, Llama's higher Spearman and Kendall Tau values indicate superior performance in maintaining relative ordering. Similarly, Mistral stands out with its higher agreement in K-Alpha and Gwet's linear coefficient, compared to models like Claude Opus3, which shows moderate performance across most metrics. These contrasting results demonstrate the potential pitfalls of relying on a single metric and reinforce our recommendation for using multiple metrics in combination, with particular emphasis on Gwet's AC2 for skewed distributions, rank correlations for ordering preservation, and task-specific metric selection based on the nature of the legal evaluation task.

\begin{tcolorbox}[colback=green!3,colframe=green!20]
\textbf{Takeaway.} When selecting LLM judges for legal RAG evaluation, avoid relying on single metrics like Krippendorff's alpha in skewed distributions. Instead, use Gwet's AC2 for agreement assessment and rank correlation coefficients for ordering consistency—this multimetric approach reveals nuanced judge capabilities that single metrics miss.
\end{tcolorbox}

\subsection{RQ2 Finding: Comparing RAG Systems}

Based on our comprehensive analysis, we recommend: (1) using non-parametric tests like the WSRT for comparing legal RAG systems due to the typically skewed nature of evaluation metrics, (2) applying B-H corrections when conducting multiple comparisons to control false discovery rates while maintaining statistical power, and (3) evaluating systems across multiple quality dimensions to capture the nuanced requirements of legal applications. These recommendations emerge from our systematic comparison of two RAG systems, where we focused on both distribution characteristics and hypothesis testing with appropriate corrections.

Our statistical analysis examined the distribution of different evaluation metrics to inform our statistical approach. As shown in Table~\ref{tab:metric_skewness}, metrics exhibited varying degrees of skewness, from slight left skewness in 'Relevance' (-0.49) to strong right skew in 'Inaccurate Hallucinations' (5.13), confirming the appropriateness of our choice of nonparametric tests.

Using the Wilcoxon signed rank test with B-H corrections for multiple comparisons, we conducted a comprehensive comparison of Systems A and B across all metrics. The results revealed distinct patterns of superiority between the systems. System B demonstrated significant advantages in relevance (adjusted p-value = 0.0358), completeness (adjusted p-value = 1.215e-18), and Correctness (adjusted p-value < 0.05). In contrast, System A showed superior performance in Extrinsic Hallucinations (adjusted p-value = 0.0204) and readability (adjusted p-value = 0.01997). Neither system showed a significant advantage in Inaccurate Hallucinations (adjusted p-values > 0.05 for both hypotheses).

These findings highlight the importance of considering multiple quality dimensions when evaluating legal RAG systems. While System B excelled in crucial accuracy-related metrics for legal reliability, System A demonstrated strengths in presentation and hallucination prevention. Balanced performance across different metrics reflects the inherent trade-offs in optimizing RAG systems for legal applications, where both factual accuracy and readability are essential for professional use.

\begin{tcolorbox}[colback=green!3,colframe=green!20]
\textbf{Takeaway.} Legal RAG systems exhibit inherent trade-offs between precision and presentation quality. The Wilcoxon Signed-Rank Test with Benjamini-Hochberg corrections provides the statistical rigor needed to detect these nuanced differences across multiple quality dimensions simultaneously.
\end{tcolorbox}
\section{Conclusion}
In conclusion, this study demonstrates the viability and challenges of using LLMs as evaluative judges for domain-specific (e.g., legal) RAG systems. Our research makes three key contributions: First, we establish that a multimetric approach to evaluating LLM judges is essential, with different metrics capturing distinct aspects of reliability. Gwet's AC2 proved particularly effective for skewed distributions common in legal evaluations, while rank correlation metrics better captured ordering relationships crucial for legal precedent. Second, we demonstrate that robust statistical analysis, particularly the Wilcoxon Signed-Rank Test with Benjamini-Hochberg corrections, is crucial for meaningful system comparisons. This approach effectively balances statistical power with false discovery control in multiple comparison scenarios. Third, our findings highlight the importance of comprehensive evaluation frameworks that consider both quantitative metrics and domain-specific requirements. While LLM judges can significantly reduce evaluation time, their effective deployment requires careful consideration of reliability metrics, statistical methods, and domain-specific constraints.

\balance
\bibliography{main}
\bibliographystyle{ACM-Reference-Format}

\section*{Appendix: Why Gwet's AC2 is More Reliable Under Skewed Labels}

Inter-rater reliability coefficients are generally defined in terms of 
the \emph{observed agreement} $A_o$ and the \emph{expected agreement} $A_e$ 
that would be obtained ``by chance.'' The generic form is:
\[
\text{Coefficient} \;=\; \frac{A_o - A_e}{\,1 - A_e\,}.
\]

\paragraph{Definitions.}
\begin{itemize}
    \item $A_o$: the empirical proportion of times that raters agree 
          (possibly weighted for ordinal data).
    \item $A_e$: the chance agreement, estimated from the overall 
          distribution of labels.
    \item $D_o = 1 - A_o$: the observed disagreement.
    \item $D_e = 1 - A_e$: the expected disagreement.
\end{itemize}

Krippendorff's $\alpha$ is commonly expressed in terms of disagreement:
\[
\alpha \;=\; 1 - \frac{D_o}{D_e}.
\]

\paragraph{Problem under skew.}
Suppose the ratings are highly imbalanced (e.g., $90\%$ of responses fall into 
a single Likert category). Then the marginal probability distribution is 
dominated by that one category, which makes $A_e \to 1$. Consequently, 
$1 - A_e \to 0$ (equivalently, $D_e \to 0$). This causes the denominator 
in both $\kappa$ and $\alpha$ to shrink toward zero, depressing the 
coefficient even when raters actually agree most of the time. 
This is the well-known \emph{prevalence paradox}.

\paragraph{How AC2 differs.}
Gwet's AC2 avoids this instability by modeling chance disagreement 
directly. Instead of relying on squared marginals, it normalizes by the 
\emph{available chance disagreement}, which is proportional to
\[
1 - \sum_{k} p_k^2,
\]
where $p_k$ is the overall proportion of ratings in category $k$. 
This term measures how much variability is present in the label 
distribution. It only vanishes when all ratings fall into a single category, 
and it decreases at the same rate as the actual difficulty of the task. 

\paragraph{Conclusion.}
Because AC2's denominator reflects the true amount of potential 
chance disagreement, it remains stable under skewed distributions. 
In contrast, $\kappa$ and $\alpha$ normalize by a vanishing term 
when one category dominates, which can yield deceptively low 
reliability scores despite high observed agreement.

\end{document}